%% file: top_program.tex
\documentclass[smallcondensed]{svjour3} 

\usepackage[round]{natbib}

\usepackage{hyphenat}
\usepackage{booktabs}
\usepackage{multirow}

\usepackage{algorithm}
\usepackage{algpseudocode}

\usepackage{amsmath}
\usepackage{amssymb}

\usepackage{graphicx}
\graphicspath{{./figures/}} 

\usepackage{subfiles}

\usepackage{subfig}

\usepackage{tabularx}

\usepackage[bookmarks=true,bookmarksdepth=3]{hyperref}

\usepackage{tablefootnote}

\newtheorem{exp-hypothesis}{Experimental Hypothesis}

\DeclareMathSymbol{\mlq}{\mathord}{operators}{``}
\DeclareMathSymbol{\mrq}{\mathord}{operators}{`'}

\makeatletter
\newcommand\newtag[2]{#1\def\@currentlabel{#1}\label{#2}}
\makeatother

\smartqed  

\begin{document}

\title{Top Program Construction and Reduction for polynomial time
Meta-Interpretive Learning}

\titlerunning{Top Program Construction and Reduction for polynomial time MIL}

\author{ S. Patsantzis \and 
	 S.H. Muggleton
}

\institute{S. Patsantzis \at
	   Imperial College London \\
	   United Kingdom \\
	   \email{ep2216@ic.ac.uk
	}
	\and
	   S. H. Muggleton \at
	   Imperial College London \\
	   United Kingdom \\
	   \email{s.muggleton@imperial.ac.uk
	}
}

\date{Received: 15 May 2020 / Accepted: date}

\maketitle

\subfile{0_abstract}
\subfile{1_introduction}
\subfile{2_related}
\subfile{3_framework}
\subfile{4_implementation}
\subfile{5_experiments}
\subfile{6_conclusions}
\subfile{7_acknowledgements}


\bibliographystyle{spbasic}
\bibliography{mybib}

\end{document}

%% file: 0_abstract.tex
\begin{abstract}
Meta-Interpretive Learners, like most ILP systems, learn by searching for a
correct hypothesis in the hypothesis space, the powerset of all constructible
clauses. We show how this exponentially-growing search can be replaced by the
construction of a Top program: the set of clauses in all correct hypotheses that
is itself a correct hypothesis. We give an algorithm for Top program
construction and show that it constructs a correct Top program in polynomial
time and from a finite number of examples. We implement our algorithm in Prolog
as the basis of a new MIL system, Louise, that constructs a Top program and then
reduces it by removing redundant clauses. We compare Louise to the
state-of-the-art search-based MIL system Metagol in experiments on grid world
navigation, graph connectedness and grammar learning datasets and find that
Louise improves on Metagol's predictive accuracy when the hypothesis space and
the target theory are both large, or when the hypothesis space does not include
a correct hypothesis because of ``classification noise" in the form of
mislabelled examples. When the hypothesis space or the target theory are small,
Louise and Metagol perform equally well.

\end{abstract}

%% file: 1_introduction.tex
\section{Introduction} 

\begin{figure}[t]
	\center
	\includegraphics[width=0.50\textwidth]{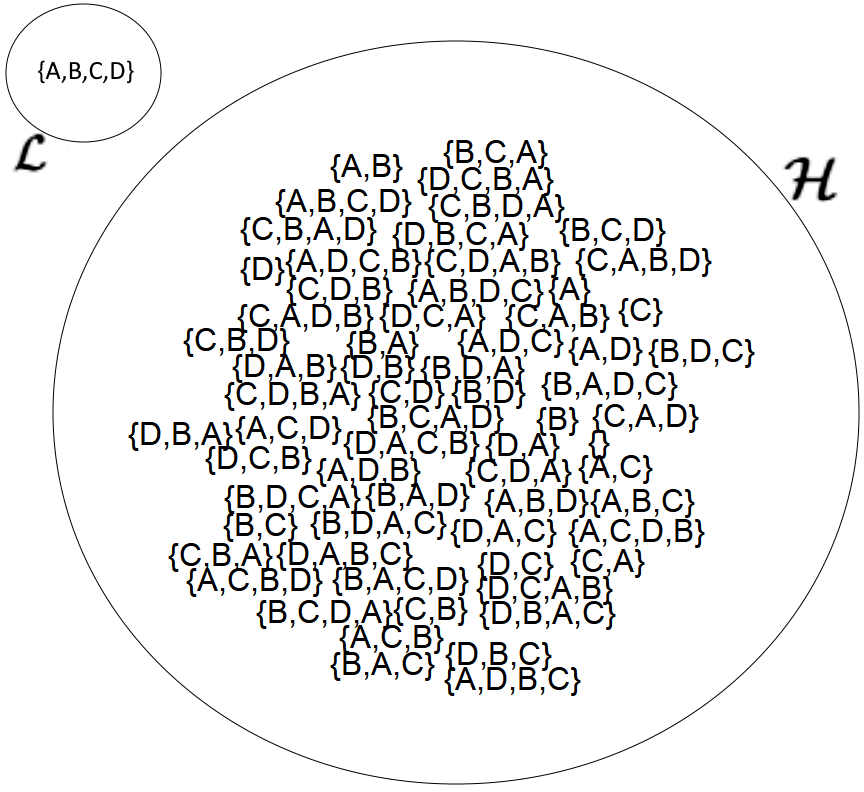}
\caption{Searching a hypothesis space $\mathcal{H}$ is an exponentially more
complex task than constructing a hypothesis language $\mathcal{L}$.}
\label{fig:motivation}
\end{figure}

Meta-Interpretive Learning (MIL) \citep{Muggleton2014} is a new setting for
Inductive Logic Programming (ILP) \citep{Muggleton1991}. ILP algorithms learn
logic theories from examples and background knowledge. MIL learners additionally
restrict the set, $\mathcal{L}$, of clauses that can be constructed from the
symbols in the background knowledge and examples (the \emph{hypothesis
language}), by means of second-order clauses called \emph{metarules}
\citep{Muggleton2014}. Each clause in $\mathcal{L}$ is an instantiation of a
metarule with existentially quantified variables substituted with predicate
symbols and constants, in a process called \emph{metasubstitution} (examples of
metarules from the MIL literature are listed in table \ref{tab:h22_metarules} in
section \ref{sec:framework}).

Like other ILP learners, the state-of-the-art MIL system, Metagol
\citep{Muggleton2014}, searches the set of hypotheses that are possible to
express as subsets of $\mathcal{L}$ for a correct hypothesis that entails all
positive examples and no negative examples. The set of hypotheses expressible in
$\mathcal{L}$ is the \emph{hypothesis space}, denoted with $\mathcal{H}$. Each
hypothesis in $\mathcal{H}$ is a set of clauses in $\mathcal{L}$, therefore
$\mathcal{H}$ is the powerset of $\mathcal{L}$ and searching $\mathcal{H}$ for a
correct hypothesis takes, in the worst case, time exponential in the cardinality
of $\mathcal{L}$.

On the other hand, enumerating the clauses in $\mathcal{L}$ need only take time
polynomial in the cardinality of $\mathcal{L}$ (see Figure
\ref{fig:motivation}). Further, the subset of $\mathcal{L}$ that includes only
the clauses in correct hypotheses in $\mathcal{H}$ \emph{is itself a correct
hypothesis}: it is the union of all correct hypotheses in $\mathcal{H}$, and,
therefore, the most general, correct set of clauses that entails each other
correct set of clauses in $\mathcal{H}$. We will call this set of clauses in
correct hypotheses the Top program and denote it by $\top$.

In the following sections we develop the framework of the Top program for MIL
and give a polynomial-time algorithm for its construction in Algorithm
\ref{alg:top_program} that is capable of learning recursive hypotheses and
performing predicate invention as described in section \ref{subsec:Future work}.
We then present a new MIL system, Louise, that implements Algorithm
\ref{alg:top_program} in Prolog and learns by Top program construction and
reduction to remove logically redundant clauses by application of Gordon
Plotkin's program reduction algorithm \citep{Plotkin1972}. Tables
\ref{tab:top_program_construction} and \ref{tab:top_program_reduction}
illustrate the inputs and outputs of Top program construction and reduction as
implemented in Louise.

\begin{table}[t]
\centering
	\begin{tabular}{l}
		\textbf{Top program construction} \\
		\toprule
		MIL problem \\
		\midrule
		$E^+ = \{ path(a,b) \leftarrow, path(a,c) \leftarrow \}$ \\
		$E^- = \{\leftarrow path(1,2), \leftarrow path(1,3) \}$ \\
		$B = \{ edge\_alpha(a,b), edge\_alpha(b,c),$ \\
			\qquad \: $edge\_alnum(a,b), edge\_alnum(b,c),$ \\
			\qquad \: $edge\_alnum(1,2), edge\_alnum(2,3) \}$ \\
		$\mathcal{M} = \{P(x,y) \leftarrow Q(x,y),$ \\
			\qquad \quad $P(x,y) \leftarrow Q(x,z), R(z,y) \}$ \\
		\midrule
		Generalisation step \\
		\midrule
		$path(x,y) \leftarrow edge\_alnum(x,y)*$ \\
		$path(x,y) \leftarrow edge\_alpha(x,y)$ \\
		$path(x,y) \leftarrow path(x,y)$ \\
		$path(x,y) \leftarrow edge\_alnum(x,z),edge\_alnum(z,y)*$ \\
		$path(x,y) \leftarrow edge\_alnum(x,z),edge\_alpha(z,y)$ \\
		$path(x,y) \leftarrow edge\_alpha(x,z),edge\_alnum(z,y)$ \\
		$path(x,y) \leftarrow edge\_alpha(x,z),edge\_alpha(z,y)$ \\
		$path(x,y) \leftarrow path(x,z),edge\_alnum(z,y)$ \\
		$path(x,y) \leftarrow path(x,z),edge\_alpha(z,y)$ \\
		\midrule
		Specialisation step \\
		\midrule
		$path(x,y) \leftarrow edge\_alpha(x,y)$ \\
		$path(x,y) \leftarrow path(x,y)$ \\
		$path(x,y) \leftarrow edge\_alnum(x,z),edge\_alpha(z,y)$ \\
		$path(x,y) \leftarrow edge\_alpha(x,z),edge\_alnum(z,y)$ \\
		$path(x,y) \leftarrow edge\_alpha(x,z),edge\_alpha(z,y)$ \\
		$path(x,y) \leftarrow path(x,z),edge\_alnum(z,y)$ \\
		$path(x,y) \leftarrow path(x,z),edge\_alpha(z,y)$ \\
		\bottomrule
	\end{tabular}
\caption{Top program construction. $E^+$: positive examples. $E^-$: negative
examples. $B$: background knowledge; $\mathcal{M}$: metarules. Clauses marked
with $*$ in the Generalisation step are removed in the Specialisation step
because they entail negative examples. The Top program is completed in the
Specialisation step.}
\label{tab:top_program_construction}
\end{table}

\begin{table}[t]
\centering
	\begin{tabular}{ll}
		\textbf{Top program reduction} \\
		\toprule
		$path(x,y) \leftarrow edge\_alpha(x,y)$ \\
		$path(x,y) \leftarrow edge\_alnum(x,z),edge\_alpha(z,y)$ \\
		$path(x,y) \leftarrow path(x,z),edge\_alnum(z,y)$ \\
		$path(x,y) \leftarrow path(x,z),edge\_alpha(z,y)$ \\
		\bottomrule
	\end{tabular}
\caption{Reduction of the Top program in table
\ref{tab:top_program_construction} by Plotkin's program reduction algorithm
(Algorithm \ref{alg:plotkins_program_reduction}).}
\label{tab:top_program_reduction}
\end{table}

\paragraph{Contributions} In this paper, we make the following contributions:
\begin{itemize}
\item Proof that the Top program is a correct hypothesis.
\item An algorithm for Top program construction.
\item Proofs that our algorithm constructs a correct Top program from a finite
	number of examples in polynomial time.
\item Louise, a new system for MIL by Top program construction and reduction.
\item Empirical comparison of Louise to the state-of-the-art MIL system,
	Metagol.
\end{itemize}

\paragraph{Structure} In section \ref{sec:related} we place our work in the
context of the ILP and MIL literature. In section \ref{sec:framework} we
describe our Top program construction algorithm and prove its correctness,
convergence and polynomial time complexity. In section \ref{sec:Implementation}
we describe Louise. In section \ref{sec:experiments} we experimentally compare
Louise to Metagol. We conclude in section \ref{sec:conclusions} with a summary
of our findings and proposed future work.

%% file: 2_related.tex
\section{Related work}
\label{sec:related}

The cardinality of $\mathcal{H}$ for MIL is upper-bounded by an exponential
function of the size of the target theory, $\Theta$, \citep{Lin2014} and when
the true cardinality of $\mathcal{H}$ approaches this upper bound, a classical
search of $\mathcal{H}$ becomes computationally infeasible on modern hardware.
As a result most single-predicate programs learned by Metagol as reported in the
MIL literature have at most 5 clauses. See e.g.
\citep{Muggleton2014,Lin2014,Muggleton2015,Cropper2015,Cropper2015DataTrans,Cropper2016,Muggleton2018,Morel2019}.

Much of the MIL literature is preoccupied with reducing the size of $\Theta$ as
a means of reducing the maximum size of $\mathcal{H}$ and thereby the cost of a
search for a correct hypothesis. In the \emph{Episodic learning}
\citep{Muggleton2014} and \emph{Dependent learning} \citep{Lin2014} settings,
Metagol learns larger multi-predicate programs by incrementally learning small
sub-programs while the variant Metagol$_{AI}$ learns from abstractions and
higher-order background knowledge \citep{Cropper2016}. Such techniques take
advantage of the theory reformulation \citep{Stahl1993} aspect of predicate
invention to reduce the size of $\Theta$ to fewer than 5 clauses and allow
learning to proceed when the complexity of a search of $\mathcal{H}$ would
otherwise be overwhelming. Top program construction is efficient when $\Theta$
is large and when $\mathcal{H}$ is large and does not require predicate
invention for the purpose of learning programs larger than 5 clauses.

Metarules, central to MIL, where originally proposed in \citep{Emde83}, where
the metarules named \emph{Chain}, \emph{Inverse} and \emph{Identity} in table
\ref{tab:h22_metarules}, representing, respectively, the concepts of
transitivity, reflexivity and symmetry between binary relations formed the basis
of a mechanism for concept discovery. This approach was further developed in
systems like METAXA.3 \citep{Emde87}, BLIP \citep{Wrobel1988} and MOBAL
\citep{Morik1993,Kietz92}.

The Top program construction procedure described in Algorithm
\ref{alg:top_program} can be contrasted to the \emph{Rule Discovery Tool} (RDT)
in MOBAL. RDT employs a generate-and-test algorithm that conducts a
general-to-specific search for a hypothesis that satisfies a user-defined
criterion, guided by a subsumption order over metarules. By contrast, Algorithm
\ref{alg:top_program} does not conduct a search and is not a generate-and-test
procedure, but a resolution-based proof procedure that restricts the set of
constructed clauses by means of the positive examples then further refines this
set by means of the negative examples. Unlike RDT, Algorithm
\ref{alg:top_program} can construct recursive hypotheses, including
left-recursive and mutually recursive ones as discussed in section
\ref{sec:Implementation}.


Other ILP systems using metarules (also called program \emph{schemata}) have
been proposed for the specific purpose of learning recursive logic programs,
like CRUSTACEAN \citep{Aha1994}, CILP \citep{Lapointe1993}, Force2
\citep{Marcinkowski1992}, Sieres \citep{Wirth1992}, TIM \citep{Idestam1996},
Synapse \citep{Flener1993}, Dialogs \citep{Flener1997} and MetaInduce
\citep{Hamfelt1994}. Such systems learn by a subsumption-order search of
$\mathcal{H}$ and are typically limited to recursive programs of restricted form
(e.g. exactly one base case and one recursive clause), or require additional
inductive biases, only accept examples of one target predicate at a time, cannot
use background knowledge or require ground background knowledge, cannot perform
predicate invention etc. \citep{Flener1999}. More recent systems ILASP,
\citep{Law2014}, that learns Answer Set Programs (but does not use metarules)
and $\delta$ILP \citep{Evans2018}, a deep neural network-based system that uses
metarules, can learn recursive programs but can perform no, or only limited,
predicate invention. Algorithm \ref{alg:top_program} can construct arbitrary
recursive hypotheses without restriction on the number of clauses, target
predicates or background knowledge and can perform predicate invention in the
\emph{Dynamic Learning} setting as discussed in section \ref{subsec:Future
work}.

A \emph{Top theory} is used by some ILP systems as e.g. in TopLog
\citep{Toplog2008} and MC-TopLog \citep{MCToplog2012} and in the non-monotonic
setting in the ASP-learning systems TAL \citep{Corapi2010InductiveLP}, ASPAL
\citep{Corapi2012} and RASPAL \citep{Athakravi2014}. A Top theory is an instance
of strong inductive bias used to direct the search of $\mathcal{H}$ which
remains expensive and which Top program construction avoids altogether.

The Top program is a unique object in $\mathcal{H}$ that can be constructed
without an expensive search. It is comparable to Least General Generalisation
(LGG) \citep{Plotkin1970, Plotkin1971}, or the Bottom clause
\citep{Muggleton1995}, also unique, directly constructible objects. The Top
program differs to the LGG and Bottom clause in that it is not a clause but a
correct hypothesis, i.e. a set of clauses.

%% file: 3_framework.tex
\section{Framework}
\label{sec:framework}

\begin{table}[t]
\centering
	\begin{tabular}{ll}
		$\boldsymbol{H_2^2}$ \textbf{metarules} & \\
		\toprule
		\emph{Abduced}    & $P(X,Y)$\\
		\emph{Identity}   & $P(x,y) \leftarrow Q(x,y)$ \\
		\emph{Inverse}    & $P(x,y) \leftarrow Q(y,x)$ \\
		\emph{Chain}      & $P(x,y) \leftarrow Q(x,z), R(z,y)$ \\
		\emph{Stack}      & $P(x,y) \leftarrow Q(x,z), R(y,z)$ \\
		\emph{Queue}      & $P(x,y) \leftarrow Q(z,x), R(z,y)$ \\
		\emph{Tailrec}    & $P(x,y) \leftarrow Q(x,z), P(z,y)$ \\
		\emph{Precon}     & $P(x,y) \leftarrow Q(x), R(x,y)$ \\
		\emph{Postcon}    & $P(x,y) \leftarrow Q(x,y), R(y)$ \\
		\bottomrule
	\end{tabular}
\caption{Examples of second-order Metarules from the MIL literature.
	As is common in the literature, quantifiers are omitted and
	quantification is instead denoted by capitalisation; $P,Q,R$:
	existentially quantified second-order variables; $X,Y$: existentially
	quantified first-order variables; $x,y,z$: universally quantified
	first-order variables.}
\label{tab:h22_metarules}
\end{table}

\subsection{Background} We follow the Logic Programming and ILP terminology
established in \citep{Nienhuys-Cheng1997} which we extend with MIL-specific
terms and terminology for second-order definite clauses and programs, as
follows.

\subsubsection{Logical notation}
\label{subsubsec:Logical notation}

$\mathcal{C}$ is the set of constants and $\mathcal{P}$ the set of predicate
symbols. First-order variables are quantified over $\mathcal{C}$ and
second-order variables are quantified over $\mathcal{P}$. An atom or literal is
second-order if it contains at least one second-order variable, or a predicate
symbol, as a term, or as an argument of a term. A definite clause is
second-order if it contains at least one second-order literal. A literal is
datalog \citep{Ceri1989} if it contains no function symbols of arity more than
0. A definite clause is datalog if it contains only datalog literals. A logic
program is definite datalog if it contains only definite datalog clauses.

\subsubsection{Meta-Interpretive Learning}
\label{subsubsec:Meta-Interpretive Learning}

MIL is a form of ILP where the first-order language of hypotheses, $\mathcal{L}$
(a set of \emph{clauses}), is defined by a set of metarules, second-order
definite clauses with existentially quantified variables in the place of
predicate symbols and constants.

The $H^2_2$ language of definite datalog metarules with at most two body
literals of arity at most 2 has Universal Turing Machine expressivity and is
decidable when $\mathcal{P}$ and $\mathcal{C}$ are finite \citep{Muggleton2015}.
Examples of $H^2_2$ metarules found in the MIL literature are given in table
\ref{tab:h22_metarules}.

Each clause in $\mathcal{L}$ is an instantiation of a metarule with second-order
existentially quantified variables substituted for symbols in $\mathcal{P}$ and
first-order existentially quantified variables substituted for constants in
$\mathcal{C}$. A substitution of the existentially quantified variables in a
metarule $M$ is a \emph{metasubstitution}, denoted as $\mu/M$.

A system that performs MIL is a Meta-Interpretive Learner, or MIL-learner (with
a slight abuse of abbreviation to support a natural pronunciation). A
MIL-learner is given the elements of a MIL problem and returns a hypothesis as a
solution to the MIL problem. A MIL problem is a quintuple, $\mathcal{T} =
\langle E^+,E^-,B,\mathcal{M},\mathcal{H} \rangle$ where: a) positive examples,
$E^+$, are ground definite atoms and negative examples, $E^-$, are ground Horn
goals, having the symbol and arity of one or more \emph{target predicates}; b)
the background knowledge, $B$, is a set of program clause definitions with
definite datalog heads; c) $\mathcal{M}$ is a set of metarules; and d)
$\mathcal{H}$ is the hypothesis space, a set of hypotheses. 

Each hypothesis in $\mathcal{H}$ is a set of clauses in $\mathcal{L}$. Each $H
\in \mathcal{H}$ is a definition of a target predicate in $E^+$ and may include
definitions of one or more \emph{invented predicates}, predicates other than a
target predicate and not defined in $B$. For each $H \in \mathcal{H}$, if $H
\wedge B \models E^+$ and $\forall e^- \in E^-: H \wedge B \not \models e^-$,
then $H$ is a correct hypothesis.

Typically a MIL learner is not explicitly given $\mathcal{H}$ or $\mathcal{L}$,
rather those are implicitly defined by $\mathcal{M}$ and the constants
$\mathcal{C}$ and symbols $\mathcal{P}$ in $E^+, E^-,B$ and any invented
predicates. The original MIL-learner, Metagol, searches $\mathcal{H}$ for a
correct hypothesis by iterative deepening on the cardinality of hypotheses. Our
new MIL-Learner Louise does not search $\mathcal{H}$ and instead constructs, and
then reduces, the Top program for $\mathcal{T}$, the set of clauses in all
correct hypotheses in $\mathcal{H}$, defined below:

\begin{definition} Let $\mathcal{T} = \langle E^+, E^-, B, M,
	\mathcal{H}\rangle$ be a MIL problem and $\mathcal{L}$ the hypothesis
	language. $\top$ is the Top program for $\mathcal{T}$ iff for all $C \in
	\mathcal{L}$ where $\exists e^+ \in E^+: C \wedge B \models e^+$ and
	$\nexists e^- \in E^-: C \wedge B \models e^-$, $C \in \top$.
\label{def:top_program}
\end{definition}

\begin{theorem} If $\mathcal{H}$ includes a correct hypothesis, $\top$ is a correct hypothesis.%
\label{thm:top_program_correctness}
\end{theorem}

\begin{proof} Assume Theorem \ref{thm:top_program_correctness} is false and
	$\mathcal{H}$ includes a correct hypothesis. Then either $\top \wedge B
	\not \models E^+$ or $\exists e^- \in E^-: \top \wedge B \models e^-$.
	Let $H \subseteq \mathcal{L}$. Either $H \wedge B \models E^+$ or $H
	\not \subseteq \top$. If $H \wedge B \models E^+$ then $\top \wedge B
	\models E^+$. $\forall C \in \mathcal{L}$ either $\nexists e^- \in E^-:
	C \wedge B \models e^-$ or $C \not \in \top$.  Therefore $\nexists e^-
	\in E^-: \top \wedge B \models e^-$. Thus the assumption is contradicted
	and Theorem \ref{thm:top_program_correctness} holds. \qed%

\end{proof}

\subsection{Top program construction}
\label{subsec:top_program_construction}

\begin{algorithm}[tb]
\caption{Top program construction}
\label{alg:top_program}
\textbf{Input}: $\mathcal{T} = \langle E^+,E^-,B,\mathcal{M},\mathcal{H} \rangle$, a MIL problem.\\
\textbf{Output}: $\top$, the Top program for $\mathcal{T}$, as a set of
metasubstitutions $\mu/M$, where $M \in \mathcal{M}$.
\begin{algorithmic}[1]
	\Procedure{Top}{$E^+,E^-,B,\mathcal{M}$}
		\State \textbf{Initialise}: Set $\top \leftarrow \emptyset$
		\State Set $\top \leftarrow$ \textsc{Generalise}($E^+,B,\mathcal{M}, \top$)
		\State Set $\top \leftarrow$ \textsc{Specialise}($E^-,B,\mathcal{M},\top$)
		\State \textbf{Return} $\top$
	\EndProcedure
        \Statex
	\Procedure{Generalise}{$E^+,B,\mathcal{M},\top$}
		\For{$e^+ \in E^+$}
			\State Set $\top \leftarrow \top \cup \{\mu/M: M \mu \cup B \cup E^+ \models e^+\}$
		\EndFor
		\State \textbf{Return} $\top$
	\EndProcedure
	\Statex
	\Procedure{Specialise}{$E^-,B,\mathcal{M},\top$}
		\For{$e^- \in E^-$}
			\State Set $\top \leftarrow \top \setminus \{\mu/M: M \mu \cup B \cup E^+ \models e^-\}$
		\EndFor
		\State \textbf{Return} $\top$
	\EndProcedure
\end{algorithmic}
\end{algorithm}

Algorithm \ref{alg:top_program} lists our algorithm for Top program
construction. To clarify, the name of Algorithm \ref{alg:top_program} is ``Top
program construction". Section \ref{sec:Implementation} describes our Prolog
implementation of Algorithm \ref{alg:top_program} as the basis of a new MIL
system called ``Louise".

In the following sections we prove that Algorithm \ref{alg:top_program}
correctly constructs the Top program for a MIL problem in polynomial time and
after processing only a finite number of examples.

\subsection{Preliminaries}
\label{subsec:Preliminaries}

\paragraph{Finite MIL problem} In the following sections, let $\mathcal{T}_k =
\langle E^+_k, E^-_k, B_k, \mathcal{M}_k, \mathcal{H}_k \rangle$ where $k$ is
the finite maximum number of body literals in each $M \in \mathcal{M}_k$. Let
$\mathcal{C}_k$ and $\mathcal{P}_k$ be the finite sets of constants and
predicate symbols in $E^+_k, E^-_k, B_k$; and let $\mathcal{L}_k$ be the
hypothesis language of clauses constructible with $\mathcal{M}_k, \mathcal{C}_k,
\mathcal{P}_k$.

\paragraph{Target theory} For each target predicate $P \in \mathcal{T}_k$, let
$\Theta_P$, a definition of $P$, be the target theory of $P$ such that each
clause in $\Theta_P$ is an instance of a metarule $M \in \mathcal{M}_k$. For
each $P$, $B_P$ is the Herbrand base of $P$; $SS(\Theta_P) \subseteq B_P$ is the
success set of $\Theta_P \cup B_k$ restricted to atoms of $P$; and $FF(\Theta_P)
= B_P \setminus SS(\Theta_P)$ is the finite failure set of $\Theta_P$,
restricted to atoms of $P$ (i.e. the set of atoms $p$ of $P$ such that there
exists a finitely-failed resolution tree for $\Theta_P \cup B_k \cup
\{\leftarrow p\}$).

\paragraph{Subsets of $\mathcal{L}_k$} Let $\top^0_k \subseteq \mathcal{L}_k$ be
the set of clauses that entail exactly 0 positive examples in $E^+_k$ with
respect to $B_k$; let $\top^+_k \subseteq \mathcal{L}_k$ be the set of clauses
that entail at least one positive example in $E^+_k$ with respect to $B_k$; and
let $\top^-_k \subseteq \mathcal{L}_k$ be the set of clauses that entail at
least one positive example in $E^+_k$ and at least one negative example in
$E^-_k$ with respect to $B_k$. Let $\top_k$ be the Top program for
$\mathcal{T}_k$. Note that $\top^0_k \cap \top^+_k = \emptyset$, $\top^-_k
\subseteq \top^+_k$ and $\top^+_k \setminus \top^-_k = \top_k$. 

\paragraph{Inductive soundness and completeness} An inductive inference
procedure is a) inductively sound, or simply sound, when it derives no clauses
that entail one or more negative examples with respect to background knowledge,
and b) inductively complete, or simply complete, when it derives all clauses
that entail one or more positive examples with respect to background knowledge.

\subsection{Inductive soundness and completeness of Algorithm \ref{alg:top_program}}
\label{subsec:Inductive soundness and completeness of Algorithm 1}

\subsubsection{Learning in the limit}
\label{subsubsec:Learning in the limit}

\begin{lemma}  Given $E^+_k = \bigcup_{P \in \mathcal{T}_k} SS(\Theta_{P}),
	B_k,\mathcal{M}_k, \emptyset$, procedure \textsc{Generalise} in
	Algorithm \ref{alg:top_program} returns $\top_k^+$.
\label{lem:proc_generalise}
\end{lemma}

\begin{proof} Follows from the finiteness of $\mathcal{P}_k, \mathcal{C}_k$ and
	the soundness and completeness of SLD resolution for definite programs
	\citep{Nienhuys-Cheng1997}. The completeness of SLD resolution ensures
	that procedure \textsc{Generalise} will derive all clauses in
	$\mathcal{L}_k$ that entail at least one positive example in $E^+_k$ and
	the soundness of SLD resolution ensures that procedure
	\textsc{Generalise} will derive no clauses in $\mathcal{L}_k$ that do
	not entail any positive examples in $E^+_k$. \qed
\end{proof}

\begin{lemma} Given $E^-_k = \bigcup_{P \in \mathcal{T}_k} FF(\Theta_P),
	B_k,\mathcal{M}_k, \top_k^+$, Procedure \textsc{Specialise} in Algorithm
	\ref{alg:top_program} returns $\top_k^+ \setminus \top_k^- = \top_k$.
\label{lem:proc_specialise}
\end{lemma}

\begin{proof} Same as for Lemma \ref{lem:proc_generalise}. The completeness of
	SLD resolution ensures that procedure \textsc{Specialise} will derive
	all clauses in $\mathcal{L}_k$ that entail at least one negative example
	in $E^-_k$ and the soundness of SLD resolution ensures that procedure
	\textsc{Specialise} will derive no clauses in $\mathcal{L}_k$ that
	entail no negative examples in $E^-_k$. \qed
\end{proof}

\begin{theorem} Algorithm \ref{alg:top_program} is inductively sound and
	complete.
\label{thm:top_soundness_completeness}
\end{theorem}

\begin{proof} Follows directly from Lemmas \ref{lem:proc_generalise},
	\ref{lem:proc_specialise}.
\end{proof}

\subsubsection{Finite example sets}
\label{subsubsec:Finite example sets}

In this section we show that Algorithm \ref{alg:top_program} can construct
$\top_k$ from finite $E^+_k,E^-_k$.

\begin{lemma} By Lemma \ref{lem:proc_generalise}, $\top_k \subseteq$
	\textsc{Generalise}$(E^+_k,B_k,\mathcal{M}_k,\emptyset)$. This implies
	$|E^+_k| \geq |\top_k|$.
\label{lemm:minimal_positive}
\end{lemma}

\begin{proof} Assume Lemma \ref{lemm:minimal_positive} is false. In this case,
	$\top_k \subseteq$ \textsc{Generalise}$(E^+_k, B_k, \mathcal{M}_k,
	\emptyset)$ and $|E^+_k| < |\top_k|$. Then $\exists e^+ \in E^+$ such
	that in line 9 of Algorithm \ref{alg:top_program}, the set $\{\mu/M : M
	\mu \cup B_k \cup E^+_k \models e^+\}$ $= \emptyset$. In this case,
	$\top_k \not \models e^+$, which contradicts Theorem
	\ref{thm:top_program_correctness}. Therefore the assumption is false and
	Lemma \ref{lemm:minimal_positive} holds. 
\qed \end{proof}

\begin{lemma} By Lemma \ref{lem:proc_specialise}, $\top^+_k \setminus \top^-_k
	=$ \textsc{Specialise}$(E^-_k,B_k,\mathcal{M}_k,\top^+_k)$. This implies
	$|E^-_k| \geq |\top^-_k|$.
\label{lemm:minimal_negative}
\end{lemma}

\begin{proof} Assume Lemma \ref{lemm:minimal_negative} is false. In this case,
	$\top^+_k \setminus \top^-_k=$
	\textsc{Specialise}$(E^-_k,B_k,\mathcal{M}_k,\top^+_k)$ and $|E^-_k| <
	|\top^-_k|$. Then, $\exists e^- \in E^-_k$ such that, in line 15 of
	Algorithm \ref{alg:top_program}, the set $\{\mu/M : M \mu \cup B_k \cup
	E^+_k \models e^-\} = \emptyset$. In this case, either $\top_k \models
	e^-$, which contradicts Theorem \ref{thm:top_program_correctness}, or
	$\top^-_k = \emptyset$ and $|E^-_k| \geq |\top^-_k|$. Therefore the
	assumption is false and Lemma \ref{lemm:minimal_negative} holds. \qed
\end{proof}

\begin{lemma} Algorithm 1 must process at most $|\top_k|$ positive examples
	and at most $|\mathcal{L}_k \setminus \top^0_k|-|\top_k|$ negative
	examples before constructing $\top_k$.
\label{lem:minimal_sets}
\end{lemma}

\begin{proof} Follows directly from Lemmas \ref{lemm:minimal_positive},
	\ref{lemm:minimal_negative}. Note that $\top^-_k = (\mathcal{L}_k
	\setminus \top^0_k) \setminus \top_k$. \qed
\end{proof}

We do not know how to exactly calculate the cardinality of $\top_k$, however in
the worst case $\top_k = \mathcal{L}_k$. It is possible to place a finite upper
bound on the cardinality of $\mathcal{L}_k$ and therefore, $\top_k$, as follows.

\begin{lemma} The cardinalities of $\mathcal{L}_k, \top_k$ are finite.
\label{lem:top_program_finiteness}
\end{lemma}

\begin{proof} $\mathcal{L}_k$ is the set of clauses constructible with $p =
	|\mathcal{P}_k|$ predicate symbols and $m = |\mathcal{M}_k|$ metarules
	of at most $k$ body literals. The cardinality of this set is at most
	$mp^{k+1}$ \citep{Cropper2018}. This number is finite because $p,m,k$
	are finite. $\top_k \subseteq \mathcal{L}_k$ therefore $|\top_k| \leq
	|\mathcal{L}_k|$ and so $|\top_k|$ is finite. \qed
\end{proof}

\begin{theorem} Algorithm \ref{alg:top_program} constructs $\top_k$ after
	processing a finite number of positive and negative examples.
\label{thm:finite_learnability}
\end{theorem}

\begin{proof} Follows directly from Lemmas \ref{lem:minimal_sets},
	\ref{lem:top_program_finiteness}.
\end{proof}

\subsection{Time complexity of Algorithm \ref{alg:top_program}}
\label{subsec:Time complexity of Algorithm 1}

In this section we show that the time complexity of Algorithm
\ref{alg:top_program} is polynomial.

\begin{theorem} The time complexity of Algorithm \ref{alg:top_program} is a
	polynomial function of $|\mathcal{L}_k|$.
\label{thm:complexity}
\end{theorem}

\begin{proof} Let $c = |E^+_k|$. The worst case for the time complexity of
	Algorithm \ref{alg:top_program} is when $\top_k = \mathcal{L}_k$ and
	each clause in $\top_k$ entails each positive example in $E^+_k$ (and 0
	examples in $E^-_k$). This is the worst case because in that case,
	procedure \textsc{Generalise} in Algorithm \ref{alg:top_program} derives
	\emph{all} clauses in $\mathcal{L}_k$ from \emph{each} example in
	$E^+_k$, i.e.  the maximum number of computations is performed for each
	example in $E^+_k$. The time complexity of Algorithm
	\ref{alg:top_program} is $O(c|\mathcal{L}_k|)$ or $O(cmp^{k+1})$. \qed
\end{proof}

\begin{remark} The number of hypotheses of at most $n$ clauses in
	$\mathcal{H}_k$ is $(m p^{k+1})^n$ \citep{Cropper2018}. Therefore, the
	time complexity of a classical search of $\mathcal{H}_k$, as in Metagol,
	is $O((cmp^{k+1})^n)$ i.e. exponential in $|\mathcal{L}_k|$.
\label{remark:complexity}
\end{remark}

%% file: 4_implementation.tex
\section{Implementation}
\label{sec:Implementation}

\begin{algorithm}[t]
	\textbf{Given}: A MIL Problem, $\mathcal{T} = \langle E^+,E^-,B,\mathcal{M},\mathcal{H} \rangle $.\\
	\textbf{Return}: A solution of $\mathcal{T}$.
	\begin{algorithmic}[1]
		\State Encapsulate $\mathcal{T}: \mathcal{T}_e = \{e(E^+),e(E^-),e(B),e(\mathcal{M})\}$. \label{step:learn_encapsulation}
		\State Set $\top_e \leftarrow \textsc{Top}(\mathcal{T}_e)$. \label{step:learn_construction}
		\State Reduce $\mathcal{M}.\top_e \cup \mathcal{T}_e \setminus e(E^-)$ yielding $\top_r$. \label{step:learn_reduction}
		\State Return $x(\top_r)$ (the excapsulation of $T_r$). \label{step:learn_excapsulation}
	\end{algorithmic}
\caption{Louise}
\label{alg:learning_procedure}
\end{algorithm}

In this section we present a new MIL-learner, Louise \citep{Louise}, written in
Prolog, that learns by Top program construction and reduction\footnote{Louise
was created alongside a new version of Metagol called Thelma, an acronym for
\emph{Theory Learning Machine}. Louise was named as a play on words with Thelma,
referencing \emph{Thelma and Louise} \citep{Scott1991}.}.

\subsection{Louise's learning procedure}
\label{subsec:Louise's learning procedure}

Louise's learning procedure is outlined in Algorithm
\ref{alg:learning_procedure}. Line numbers listed in this section refer to the
numbered lines in the listing of Algorithm \ref{alg:learning_procedure}.

\begin{table}[t]
	\begin{tabular}{ll}
		\textbf{Atom or clause} & \textbf{Encapsulation}\\
		\toprule
		$edge(a,b)$ & $m(edge,a,b)$ \\
		$P(x,y) \leftarrow Q(x,z), R(z,y)$ & $m(P,x,y) \leftarrow m(Q,x,z), m(R,z,y)$ \\
		$path(x,y) \leftarrow edge(x,z), edge(z,y)$ & $m(path,x,y) \leftarrow m(edge,x,z), m(edge,z,y)$ \\
		\bottomrule
	\end{tabular}
\caption{Encapsulation of atoms and clauses, including metarules. The
excapsulation of an encapsulated atom or clause is the same as its
un-encapsulated form.}
\label{tab:encapsulation}
\end{table}

Learning begins with the \emph{encapsulation} of a MIL problem (line
\ref{step:learn_encapsulation}). An encapsulation $e(L)$ of a literal $L =
p(s_1,...,s_n)$ is a first-order atom $m(p,s_1,...,s_n)$ where $m$ is an
\emph{encapsulation predicate}. The symbol $m$ is chosen arbitrarily and has no
special meaning. The arity of each encapsulation predicate is $n+1$ where $n$ is
the arity of the encapsulated predicate(s). Therefore, a literal of a predicate
$p/n$ is encapsulated by a literal of $m/(n+1)$. An encapsulation $e(C)$ of a
definite clause $C = \{L_1,...,L_n\}$ is the set of encapsulations of literals
in $C$, $\{e(L_1),...,e(L_n)\}$. An encapsulation $e(\Pi)$ of a definite program
$\Pi = \{C_1,...,C_n\}$ is the set of encapsulations of clauses in $\Pi$,
$\{e(C_1),...,e(C_n)\}$. Table \ref{tab:encapsulation} illustrates encapsulation
for first order atoms and clauses, and metarules. Encapsulation of metarules
ensures the decidability of unification between metarule literals and literals
of first-order clauses \citep{Muggleton2015}. Encapsulation of a MIL problem
facilitates the efficient and simple construction of the Top program, $\top_e$
(line \ref{step:learn_construction}), by resolution as described below.

Our implementation of procedures \textsc{Generalise} and \textsc{Specialise} in
Louise unifies the encapsulation of each (positive or negative) example atom to
the encapsulated head literal of each metarule and resolves the metarule's
encapsulated body literals with $e(B)$ and $e(E^+)$. Resolution with $e(E^+)$
permits the derivation of clauses that have body literals with the symbol of a
target predicate and therefore the construction of a recursive Top program.
Because $e(E^+)$ is a set of ground atoms, each encapsulated body literal with
the symbol of a target predicate has a finite refutation sequence so recursive
clauses can be derived without resolution entering an infinite recursion. When
$e(E^+)$ includes multiple target predicates mutually recursive clauses can be
derived. Table \ref{tab:odd_even_mlp} lists an example of a Top program with
mutually recursive clauses derived from resolution with the encapsulation of the
background predicate $predecessor/2$ in $e(B)$ and the encapsulated examples of
the two target predicates, $odd/1$ and $even/1$ in $e(E^+)$.

$\top_e$, the result of resolving the body literals of encapsulated metarules
with $e(B)$ and $e(E^+)$ is a set of metasubstitutions. Metasubstitutions in
$\top_e$ are applied to the corresponding metarules (noted as
$\mathcal{M}.\top_e$ on line \ref{step:learn_reduction}) yielding a set of
encapsulated definite clauses, the encapsulated Top program.

Redundant clauses are removed from the encapsulated Top program by Algorithm
\ref{alg:plotkins_program_reduction} (line \ref{step:learn_reduction}). The set
of clauses remaining after reduction, $\top_r$, is then \emph{excapsulated} and
returned as the learned hypothesis, a definition of the target predicates in
$E^+$ (line \ref{step:learn_excapsulation}). Excapsulation is the opposite
process of encapsulation. An excapsulation, $x(e(L)) = L$ of an encapsulated
literal, $e(L) = m(p,s_1,...,s_n)$, is a first order literal $L =
p(s_1,...,s_n)$. An excapsulation, $x(e(C)) = C$, of an encapsulated clause
$e(C) = \{e(L_1),...,e(L_n)\}$, is a first order definite clause $C = \{L_1,
..., L_n\}$ where each $L_i$ is the excapsulation of a literal in $e(C)$. An
excapsulation, $x(e(\Pi)) = \Pi$ of an encapsulated program, $e(\Pi)$, is a set
of first order definite clauses $\Pi = \{C_1,...,C_n\}$ where each $C_i$ is the
excapsulation of a clause in $e(\Pi)$.

\subsection{Plotkin's program reduction}
\label{subsec:Plotkin's program reduction}

\begin{table}[t]
	\centering
	\begin{tabular}{l}
		\textbf{Even and odd}\\
		\toprule
		MIL problem \\
		\midrule
		$E^+ = \{ even(0) \leftarrow, even(s(s(0))) \leftarrow, odd(s(0)) \leftarrow, odd(s(s(s(0)))) \leftarrow \} $ \\
		$E^- = \{ \leftarrow even(s(0)), \leftarrow even(s(s(s(0)))), \leftarrow odd(0), \leftarrow odd(s(s(0))) \}$ \\
		$B = \{ predecessor(s(0),0). $\\ 
		$\qquad \; \; predecessor(s(s(x)),s(x)) \leftarrow predecessor(s(x),x). \}$ \\
		$\mathcal{M} = \{P(x) \leftarrow Q(x,y), R(y) \}$ \\
		\midrule
		Learned hypothesis \\
		\midrule
		$even(0).$ \\
		$even(x) \leftarrow predecessor(x,y),odd(y).$ \\
		$odd(x) \leftarrow predecessor(x,y),even(y).$ \\
		\bottomrule
	\end{tabular}
\caption{Multi-predicate MIL problem for $odd/1$ and $even/1$ and mutually
recursive hypotheses learned by Louise.}
\label{tab:odd_even_mlp}
\end{table}

\begin{algorithm}[t]
	\textbf{Given}: A set of clauses $H$.\\
	\textbf{Return}: A reduction, $H'$, of $H$.
	\begin{algorithmic}[1]
		\State Set $H'$ to $H$.
		\State Stop if every clause in $H'$ is marked [and return $H'$].
		\State Choose an unmarked clause $C$, in $H$.
		\State If $H' \setminus \{C\} \preceq \{C\}$ then change $H'$ to
			$H' \setminus \{C\}$. Otherwise, mark $C$.
		\State Go to (2).
	\end{algorithmic}
\caption{Reduction of a set of clauses (Gordon Plotkin)}
\label{alg:plotkins_program_reduction}
\end{algorithm}

In Algorithm \ref{alg:learning_procedure}, the Top program, $\top_e$, is reduced
by Gordon Plotkin's program reduction algorithm, described in
\citep{Plotkin1972} as Theorem 3.3.1.2, reproduced here as Algorithm
\ref{alg:plotkins_program_reduction} in Plotkin's original notation.

In Algorithm \ref{alg:plotkins_program_reduction}, $\Phi \preceq \Psi$ means
that ``$\Phi$ generalises $\Psi$". The generalisation of $\Psi$ by $\Phi$ is
considered with respect to a theorem, $Th$ (sic). In the context of Algorithm
\ref{alg:learning_procedure}, $Th$ is the union of the encapsulated $E^+,B$ and
$\mathcal{M}$ and applied Top program. In our implementation of Plotkin's
algorithm in Louise, $\Phi \preceq \Psi$ is true iff $\Psi$ can be derived from
$\Phi$ by Prolog's SLD-Resolution.

%% file: 5_experiments.tex
\section{Experiments}
\label{sec:experiments}

\begin{table}[t]
	\begin{tabularx}{\columnwidth}{lllllllX}
		\multicolumn{8}{l}{\textbf{Experiment datasets \& MIL problems}} \\
		\toprule
		& $|E^+|$ & $|E^-|$ & $|B|$ & $|\mathcal{M}|$ & $|\Theta|$ & $max |\mathcal{L}|$ & $max |\mathcal{H}|$ \\
		\midrule
		Grid world         & 625  & 0  & 19 & 5 & $|E^+|$ & 81,450,625  & 2.037104e+4944 \\
		Coloured graph (1) & 108  & 74 & 9  & 4 & 4   & 46,656      & 4.738381e+18 \\
		Coloured graph (2) & 108  & 74 & 1  & 4 & 4   & 64          & 16,777,216 \\
		M:tG fragment      & 1348 & 0  & 60 & 1 & 36  & 216,000     & 1.097324e+192 \\
		\bottomrule
	\end{tabularx}
\caption{Dataset summary. $|B|$: number of BK definitions. $\Theta$: target
theory. Grid world $\Theta$ is not known but $|E^+|$ approximates its
cardinality. $max|\mathcal{L}|$ is calculated as $|\mathcal{M}||B|^{k+1}$, where
$k$ is the number of literals in metarules: 3 for Grid world, otherwise 2.
$max|\mathcal{H}|$ is calculated as $max|\mathcal{L}|^{|\Theta|}$. See Lemma
\ref{lem:top_program_finiteness} and Remark \ref{remark:complexity}.}
\label{tab:mil_problem_summary}
\end{table}

\begin{figure}[t]
	\centering
	\subfloat[Grid world \label{fig:grid_world_results}]{\includegraphics[width=0.35\columnwidth]{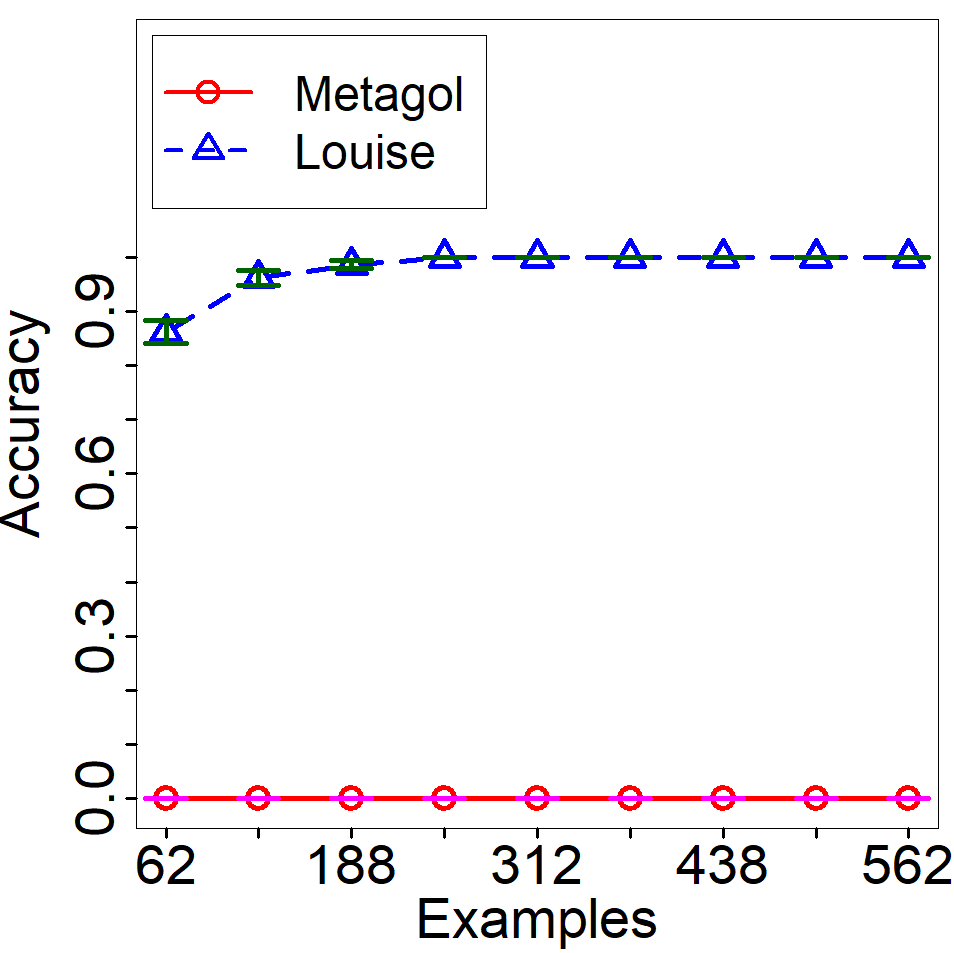}} 
	\subfloat[M:tG Fragment \label{fig:mtg_fragment_results}]{\includegraphics[width=0.35\columnwidth]{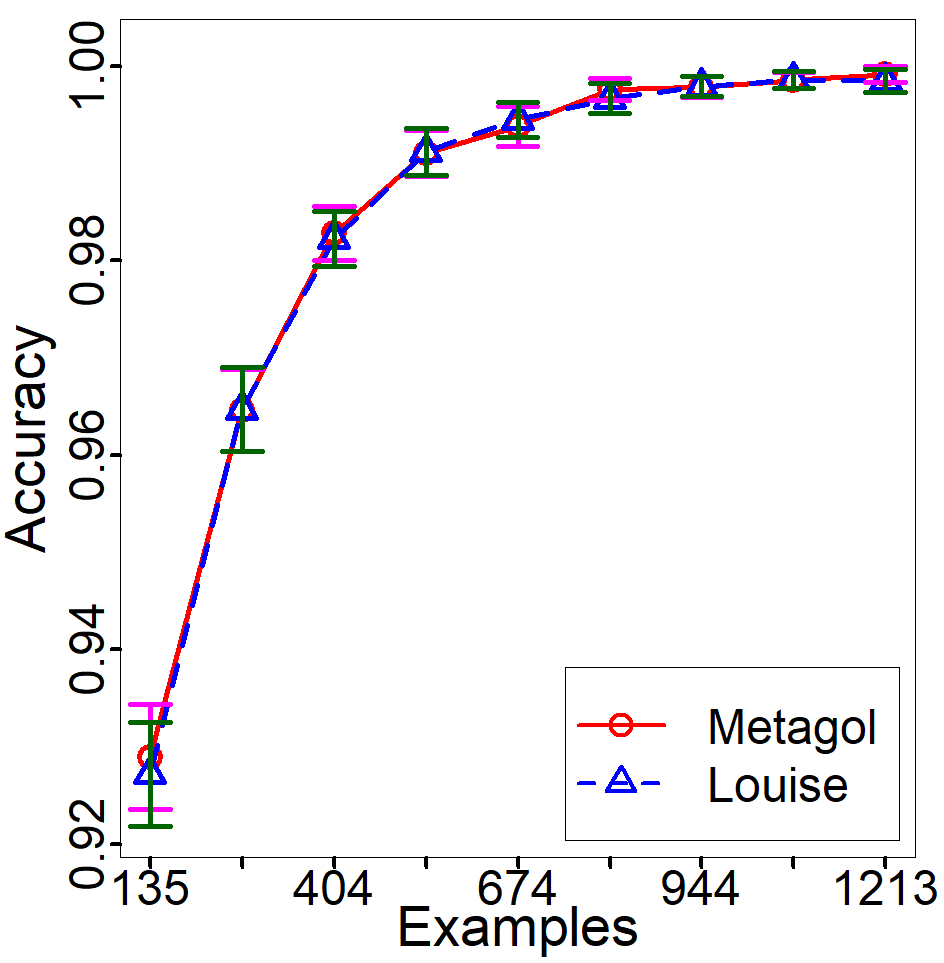}}\\ 
	\subfloat[Coloured graph (Large $\mathcal{H}$) \label{fig:redundant_bk}]{\includegraphics[width=0.5\columnwidth]{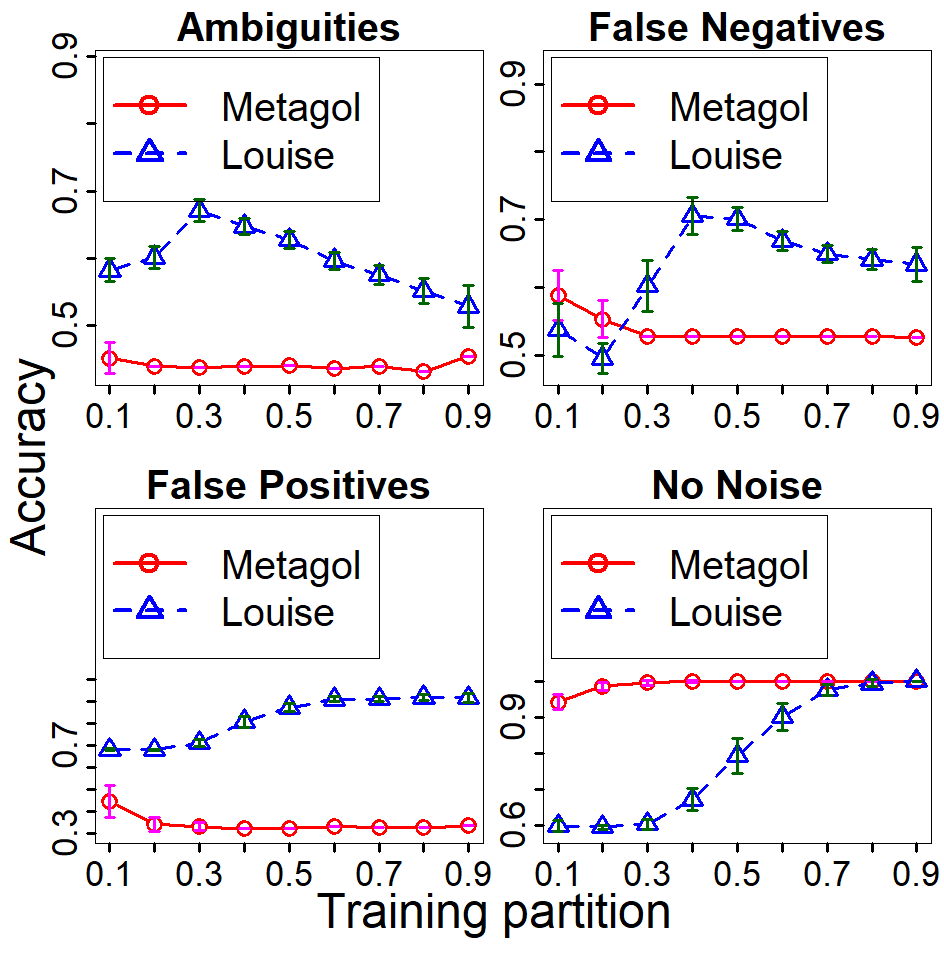}}
	\subfloat[Coloured graph (Small $\mathcal{H}$) \label{fig:no_redundant_bk}]{\includegraphics[width=0.5\columnwidth]{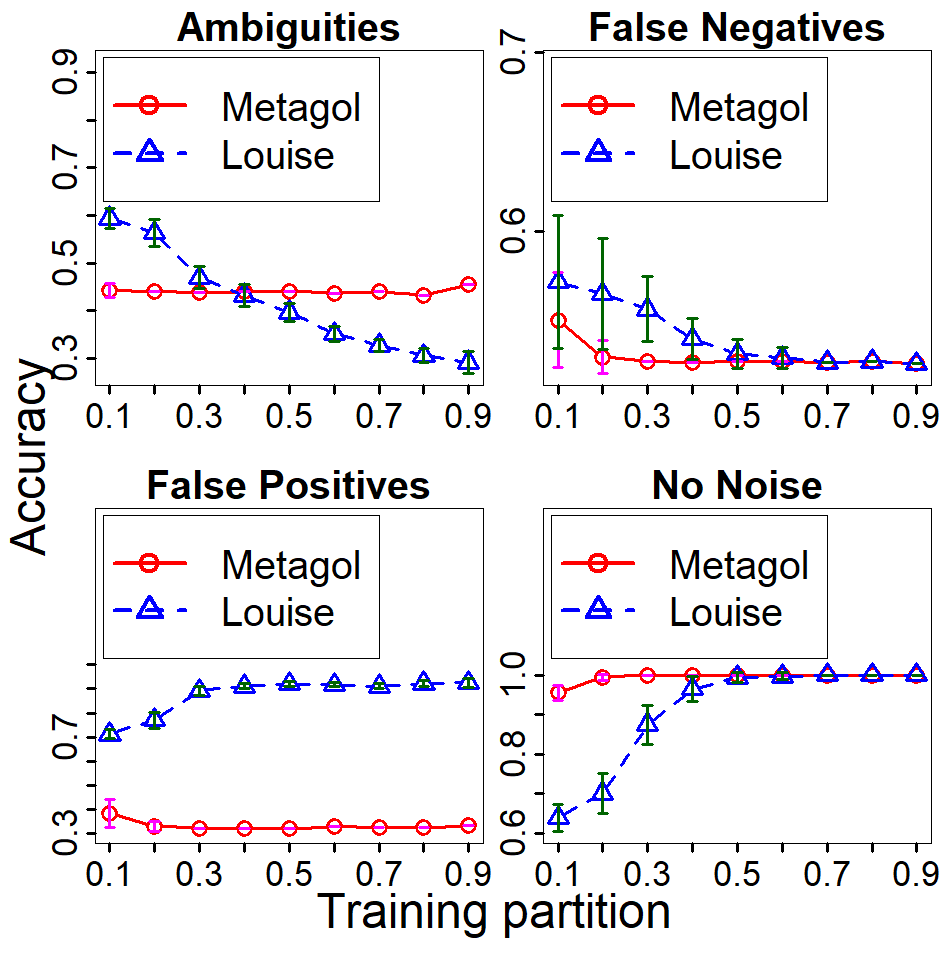}}
\caption{Learning curve experiment results (accuracy). Red circles: Metagol. Blue
triangles: Louise. x-axis: size of training partition; y-axis: accuracy on
testing partition. Error bars: standard error.}
\label{fig:experiment_results}
\end{figure}

\begin{figure}[t]
	\centering
	\subfloat[Grid world \label{fig:grid_world_time}]{\includegraphics[width=0.35\columnwidth]{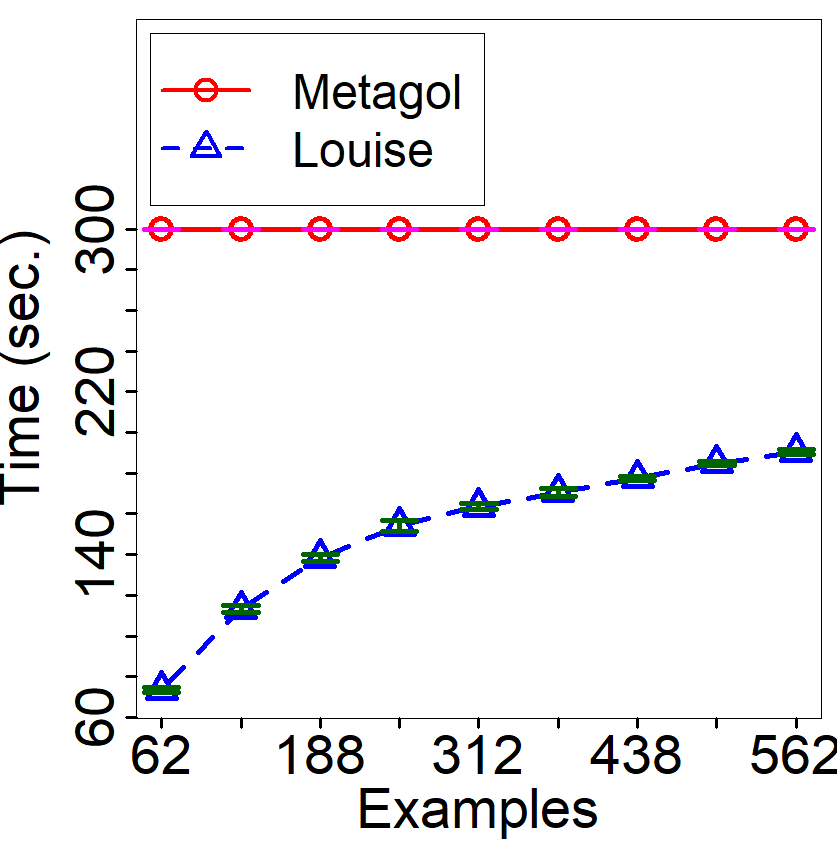}} 
	\subfloat[M:tG Fragment \label{fig:mtg_fragment_time}]{\includegraphics[width=0.35\columnwidth]{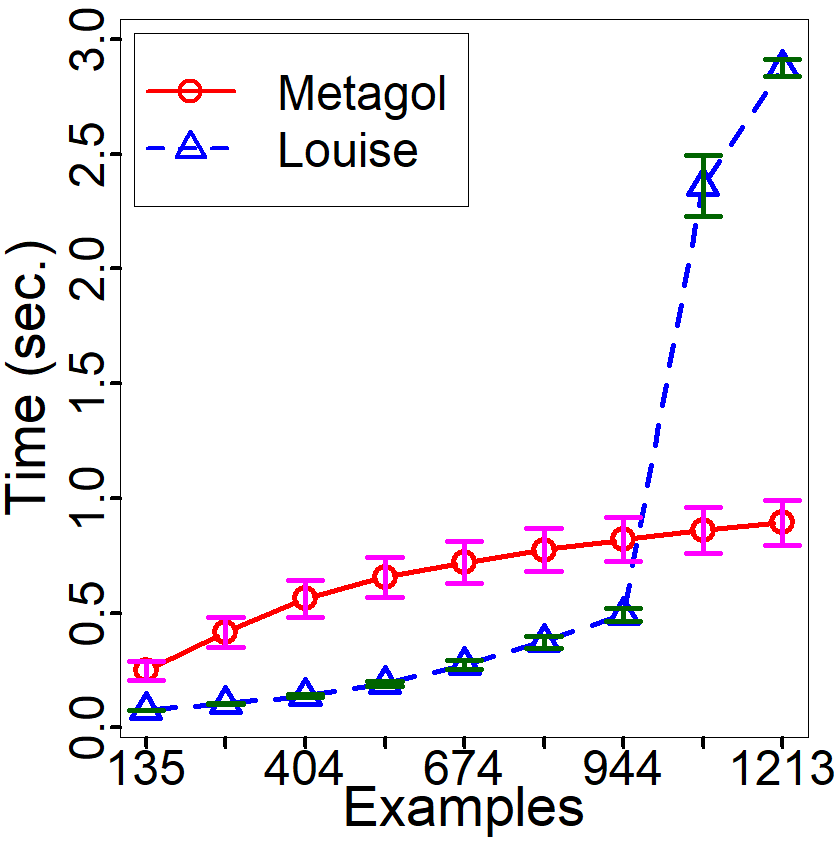}}\\ 
	\subfloat[Coloured graph (Large $\mathcal{H}$) \label{fig:redundant_bk_time}]{\includegraphics[width=0.5\columnwidth]{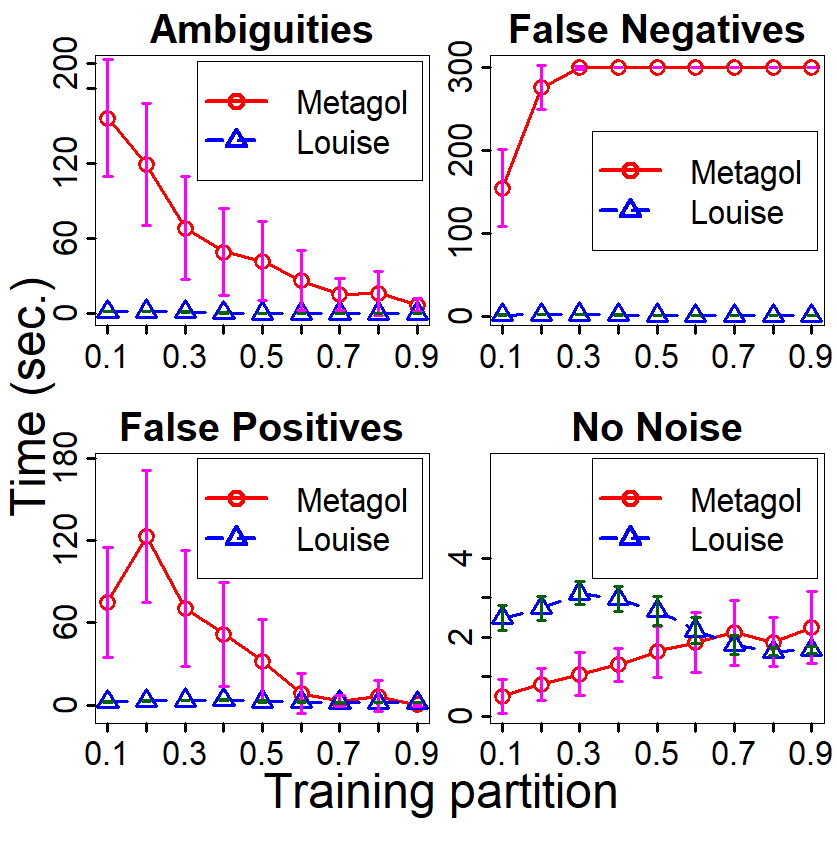}}
	\subfloat[Coloured graph (Small $\mathcal{H}$) \label{fig:no_redundant_bk_time}]{\includegraphics[width=0.5\columnwidth]{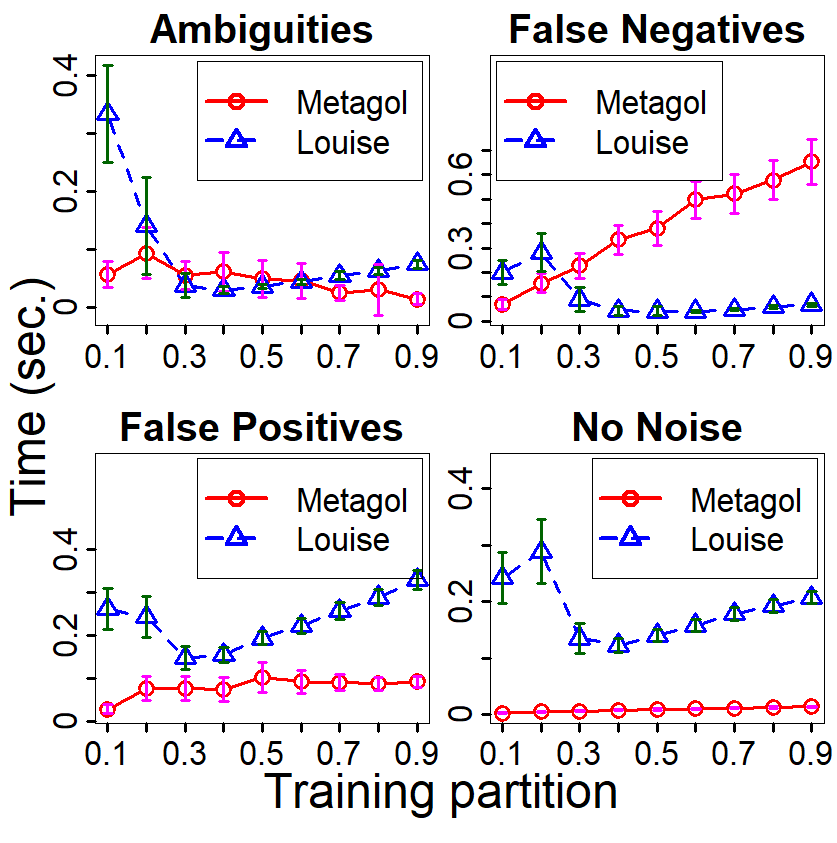}}
\caption{Learning curve experiment results (training times). Red circles:
Metagol. Blue triangles: Louise. x-axis: size of training partition; y-axis:
mean time of a training step. Error bars: standard error.}
\label{fig:experiment_timing}
\end{figure}

A MIL system that learns by Top program construction should outperform a
search-based MIL system when the complexity of a search of $\mathcal{H}$ is
maximised. Metagol's iterative deepening search orders $\mathcal{H}$ by
hypothesis size and the complexity of its search is maximised when $\mathcal{H}$
and the target theory, $\Theta$, are both large, therefore Louise should
outperform Metagol when both these conditions hold. We formalise this
expectation as Experimental Hypothesis \ref{hyp:maximal_search}:

\begin{exp-hypothesis} Louise outperforms Metagol when $\mathcal{H}$ and
	$\Theta$ are large.%
\label{hyp:maximal_search}
\end{exp-hypothesis}

When $\mathcal{H}$ does not contain a correct hypothesis, e.g. when $E^+, E^-$
have mislabelled examples (``classification noise"), a search-based MIL system
must exit with failure and its accuracy is minimal. In the worst case,
$\mathcal{H}$ is additionally large and the MIL system must perform an
exhaustive search before returning with failure. Algorithm 1 constructs as much
of $\top$ as possible given the elements of a MIL problem and so returns an
approximately correct hypothesis when a correct hypothesis does not exist. In
such situations we should expect Louise to outperform Metagol. We formalise this
expectation as Experimental Hypothesis \ref{hyp:noise}: 

\begin{exp-hypothesis} Louise outperforms Metagol when $\mathcal{H}$ does not
	include a correct hypothesis.
\label{hyp:noise}
\end{exp-hypothesis}

When $\mathcal{H}$ or $\Theta$ are small, Louise should not have an advantage
over Metagol. A special case of this is when $\mathcal{H}$ includes a single
hypothesis which is, tautologically, the set of clauses in all correct
hypotheses, i.e. the Top program. In that special case, Louise and Metagol
should perform equally well. We formalise this expectation as Experimental
Hypothesis \ref{hyp:same_accurracy}:

\begin{exp-hypothesis} Louise and Metagol perform equally when $\mathcal{H} =
	\{\top\}$.%
\label{hyp:same_accurracy}
\end{exp-hypothesis}

To test these three experimental hypotheses we compare Metagol and Louise on a
real-world dataset and two synthetic datasets summarised in table
\ref{tab:mil_problem_summary}. The synthetic Coloured graph dataset can be
configured to include ``noise" in the form of mislabelled examples and has two
variants with a small and large $\mathcal{H}$, marked with (1) and (2)
respectively in table \ref{tab:mil_problem_summary}.

\subsection{Experiment setup}
\label{subsec:Experiment setup}

We compare Metagol and Louise in a series of ``learning curve" experiments,
where we vary the number of training examples and measure predictive accuracy
and training time.

Each learning curve experiment proceeds for $k = 100$ steps. In each step we
sample, at random and without replacement, a proportion, $s$, of $E^+, E^-$ to
form a training partition. Remaining examples form the testing partition. $S$ is
taken from the sequence: $S = \langle 0.1,0.2,0.3,0.4,0.5,0.6,0.7,0.8,0.9
\rangle$. At each step, we train each learner on the training partition and
measure the accuracy of the returned hypothesis on the testing partition and the
duration of training in seconds. We set a time limit of 300 sec. for each
training step. If a training step exhausts this time limit, we calculate
the accuracy of the empty hypothesis on the testing partition. Finally, we
return the mean and standard error of the accuracy and duration for the same
sampling ratio $s$ at each step\footnote{Experiment code and datasets are
available from: $https://github.com/stassa/mlj\_2020$}.

All experiments were run on a PC with 32 8-core Intel Xeon E5-2650 v2 CPUs
clocked at 2.60GHz, with 251 Gb of RAM, running Ubuntu 16.04.6. Running each
instance of the learning curve experiment (one instance per dataset) occupied
one core of the machine at 100\% of capacity (experiments were run in parallel
as background linux jobs). The longest-running experiment was on the Coloured
Graph with False Negatives dataset with large $\mathcal{H}$ (described in
section \ref{subsec:Experiment 2: Coloured graph}) and took three days for
Metagol (but only a few hours for Louise) to complete. The shortest-running
experiment was on the M:tG Fragment dataset (described in section
\ref{subsec:Experiment 3: M:tG Fragment}) and took both systems about 11 minutes
to complete. Other experiments were completed in about 8 hours on average.


\subsection{A note on metarule selection}
\label{subsec:A note on metarule selection}

In MIL practice, metarules are typically selected manually, according to user
intuition or domain knowledge, although minimal sets of metarules for language
fragments such as $H^2_2$ (see section \ref{subsubsec:Meta-Interpretive
Learning}) have been identified, e.g. in \citep{Cropper2015,Cropper2018}. For
the experiments described in the following sections, we have manually selected
metarules as follows. 

For the Coloured Graph (section \ref{subsec:Experiment 2: Coloured graph}) and
M:tG Fragment (section \ref{subsec:Experiment 3: M:tG Fragment}) datasets where
$\Theta$ was known, we extracted metarules from the clauses of $\Theta$ with
Louise's \emph{metarule extraction} module. This defines Prolog predicates to
``lift" sets of program clauses to the second order, by variabilisation of their
predicate symbols and constants, and encapsulate them as metarules\footnote{When
$\Theta$ is not known, it is sometimes useful to extract metarules from $B$.}.

For the Grid World dataset in section \ref{subsec:Experiment 1: Grid world},
were $\Theta$ was not known, we initially selected the \emph{Chain} metarule
(table \ref{tab:h22_metarules}), that represents transitivity, such as the
relation between consecutive moves over contiguous ``cells" in a grid world,
reflecting our intuition about the likely structure of $\Theta$. Algorithm
\ref{alg:top_program} can construct recursive instances of metarules without
restriction, but Metagol imposes a \emph{lexicographic} ordering on the
predicate symbols in metasubstitutions \citep{Muggleton2015} which precludes
recursive instances of \emph{Chain} and in general requires recursive metarules
to be specified explicitly. Adding one metarule for each recursive variant of
\emph{Chain} would increase the size of $\mathcal{H}$ and penalise Metagol's
time complexity; but omitting any recursive metarules would penalise the
expressivity of $\mathcal{L}$ only for Metagol. We elected to add the
tail-recursive version of \emph{Chain}, \emph{Tailrec} (table
\ref{tab:h22_metarules}), as the only explicitly recursive metarule, by way of a
compromise. Finally, we defined three variants of \emph{Chain}, listed in table
\ref{tab:robot_dataset}, each with one or two body literals of arity 3, to allow
the use of higher-order moves defined as arity-3 predicates. 

\subsection{Experiment 1: Grid world} 
\label{subsec:Experiment 1: Grid world} 

\begin{table}[t]
	\centering
	\begin{tabular}{l}
		\textbf{Grid world} \\
		\toprule
		Navigation tasks (positive examples) \\
		\midrule
		$move([0/0,0/0,4 \times 4],[0/0,0/0,4 \times 4]).$ $move([0/0,0/2,4 \times 4],[0/2,0/2,4 \times 4]).$ \\
		$move([0/0,0/1,4 \times 4],[0/1,0/1,4 \times 4]).$ $move([0/0,0/3,4 \times 4],[0/3,0/3,4 \times 4]).$ \\
		\midrule
		Primitive moves \\
		\midrule
		$move\_up([0/0,G,4 \times 4],[0/1,G,4 \times 4]).$ \quad \! $move\_right([0/0,G,4 \times 4],[1/0,G,4 \times 4]).$ \\
		$move\_down([0/1,G,4 \times 4],[0/0,G,4 \times 4]).$ $move\_left([1/0,G,4 \times 4],[0/0,G,4 \times 4]).$  \\
		\midrule
		Composite moves \\
		\midrule
		$move\_right\_twice(Ss,Gs) \leftarrow move\_right(Ss,Ss_1),move\_right(Ss_1,Gs).$ \\
		$move\_right\_then\_up(Ss,Gs) \leftarrow move\_right(Ss,Ss_1),move\_up(Ss_1,Gs).$ \\
		\midrule
		Higher - order moves \\
		\midrule
		$double\_move(M,Ss,Gs) \leftarrow move(M),call(M,Ss,Ss_1),call(M,Ss_1,Gs).$ \\
		$triple\_move(M,Ss,Gs) \leftarrow move(M) ,call(M,Ss,Ss_1), double\_move(M,Ss_1,Gs).$ \\
		\midrule
		Triadic \emph{Chain} variants \\
		\midrule
		\emph{Tri-Chain 1}: $P(x,y)\leftarrow Q(M,x,z), R(z,y)$ \\
		\emph{Tri-Chain 2}: $P(x,y)\leftarrow Q(x,z), R(M,z,y)$ \\
		\emph{Tri-Chain 3}: $P(x,y)\leftarrow Q(M_1,x,z), R(M_2,z,y)$ \\
		\bottomrule
	\end{tabular}
\caption{Grid world dataset. In navigation tasks and primitive moves each list
of the form $[R,G,W  \times  H]$ is a grid world-state listing the location of
the agent ($R$), its goal ($G$) and the world dimensions $W \times H$. In
primitive moves, $G$ is a variable binding to the coordinates of the task's
goal (which remains unchanged during a move). In composite and higher-order moves
$Ss$ and $Gs$ are variables binding to the world states at the start and end of a
move, respectively. In higher-order moves the literal $move(M)$
nondeterministically generates the predicate symbols of primitive and composite
moves. In variants of \emph{Chain}, existentially quantified variables $\{Q,R\}$
of literals with arity 3 can only take values from the set of predicate symbols
of higher-order moves that also have arity 3, whereas existentially quantified
variables $\{M,M_1,M_2\}$ of literals with arity 2 can only take values from the
set of symbols of primitive and composite moves that have arity 2. For example,
in the first body literal in \emph{Tri-Chain 1}, a possible metasubstitution is
$\{Q/double\_move,M/move\_down\}$ resulting in a literal
$double\_move(move\_down,x,z)$ i.e. a double-move downwards.}
\label{tab:robot_dataset}

\end{table}

We create a generator for navigation problems where an agent must move to a goal
location on an empty grid world represented as a Cartesian plane with the origin
at $(0,0)$ and extending to a point $(w,h)$. Our generator takes as parameters
the $w,h$ dimensions of the grid world and generates a) all navigation tasks
between pairs of locations in the grid world as atoms of the target predicate,
$move/2$ and b) a set of primitive moves that move the agent up, down, left or
right. We define a set of composite moves that each combine two primitive moves
and two higher-order moves that repeat a primitive or composite move twice or
thrice. To form a MIL problem for this dataset we give all $move/2$ atoms as
positive examples, all primitive, composite and higher-order moves as background
knowledge and as metarules \emph{Chain} and \emph{Tailrec} from Table
\ref{tab:h22_metarules}, and three arity-3 variants of \emph{Chain} necessary
for the use of higher-order moves. No navigation task is impossible on an empty
grid world, therefore there are no negative examples. Table
\ref{tab:robot_dataset} illustrates the elements of the MIL problem.

We do not know the target theory for this problem but in preliminary experiments
Louise learns a hypothesis of 2,567 clauses from all examples and Metagol a
hypothesis equal in size to a small training sample of 5 examples, indicating a
large $\mathcal{H}$ and $\Theta$. We run our experiment in a $4 \times 4$ world
for only 10 steps after Metagol runs for more than a day when trained on 6
examples in a larger world.

\subsubsection{Grid world - results} \label{subsubsec:Grid world - results}

Figures \ref{fig:grid_world_results} and \ref{fig:grid_world_time} plot the
accuracy and training time results of the Grid world experiment, respectively.
Louise quickly learns a correct hypothesis that generalises well on the testing
partition whereas Metagol exhausts the training time limit of 300 sec. early in
the experiment, when the training partition includes only 62 examples. This
confirms Experimental Hypothesis \ref{hyp:maximal_search}.

\subsection{Experiment 2: Coloured graph} 
\label{subsec:Experiment 2: Coloured graph} 

\begin{table}[t]
	\centering
	\begin{tabular}{l}
		\textbf{Coloured Graph: MIL problem} \\
		\toprule
		Target theory \\
		\midrule
		$connected(x,y)\leftarrow ancestor(x,y).$ \\
		$connected(x,y)\leftarrow ancestor(y,x).$ \\
		$connected(x,y)\leftarrow ancestor(z,x), ancestor(z,y).$ \\
		$connected(x,y)\leftarrow ancestor(x,z), ancestor(y,z).$ \\
		\toprule
		Background knowledge \\
		\midrule
		$ancestor(x,y)\leftarrow parent(x,y). $ \\
		$ancestor(x,y)\leftarrow parent(x,z),ancestor(z,y).$ \\
		\midrule
		$parent(x,y)\leftarrow blue\_parent(x,y).$ \\
		$parent(x,y)\leftarrow red\_parent(x,y).$ \\
		\midrule
		$blue\_parent(a,c).$ \! \quad $red\_parent(k,c).$ \! \quad $blue(a).$ \quad $red(i).$ \\
		$blue\_parent(a,n).$ \quad $red\_parent(k,n).$ \quad $blue(b).$ \quad $red(j).$ \\
		$blue\_parent(b,i).$ \! \quad $red\_parent(l,i).$ \, \quad $blue(c).$ \quad $red(k).$ \\
		\bottomrule
	\end{tabular}
\caption{Target theory and (partial) BK definitions of Coloured graph datasets.}
\label{tab:coloured_graph_dataset}
\end{table}

\begin{table}[t]
	\centering
	\begin{tabular}{lll}
		\multicolumn{3}{l}{\textbf{Coloured Graph: mislabelled examples}} \\
		\toprule
				& $E^+$ & $E^-$ \\
		\midrule
		No Noise	& $E^+$ 		& $E^-$ \\
		Ambiguities	& $E^+ \cup E^-_m$ 	& $E^- \cup E^+_m$ \\
		False positives	& $E^+ \cup E^-_m$ 	& $E^- \setminus E^-_m$ \\
		False negatives	& $E^+ \setminus E^+_m$ & $E^- \cup E^+_m$ \\
		\bottomrule
	\end{tabular}
\caption{Composition of positive and negative example sets in Coloured graph
datasets. $E^+_m \subseteq E^+$ and $E^-_m \subseteq E^-$ are sets of
``mislabelled" examples selected at random and without replacement. Examples
are mislabelled by including them in the opposite set of examples. For the
Ambiguities dataset, mislabelled examples are included in both $E^+$ and $E^-$.
For the false positive and false negative examples, mislabelled examples are
removed from one and added to the other set.}
\label{tab:coloured_graph_composition}
\end{table}

To test Experimental Hypothesis \ref{hyp:noise} we create a generator for MIL
problems with a definition of the predicate $connected/2$, illustrated in table
\ref{tab:coloured_graph_dataset}, as a target theory, representing the
connectedness relation on a directed, acyclic, two-colour graph. Our generator
can produce three datasets with different kinds of mislabelled examples:
\emph{False Positives} (with negative examples mislabelled as positive),
\emph{False Negatives} (with positive examples mislabelled as negative) and
\emph{Ambiguities} (with examples simultaneously labelled positive and
negative). A fourth dataset, \emph{No Noise} is noise-free. We ``label" examples
as positive or negative by inclusion in $E^+$ or $E^-$, respectively. Table
\ref{tab:coloured_graph_composition} outlines the mislabelling process.

To create a MIL problem for each dataset we begin by generating all positive and
negative atoms of $connected/2$ forming the initial $E^+, E^-$. We select a
proportion $N$ of each set of examples at random and without replacement and
mislabel them as described above. $N = 0.2$ for each ``noisy" dataset and $N =
0$ for the No Noise dataset. We give as background knowledge the definitions of
the three arity-2 predicates used to define the target theory, $ancestor/2$,
$red\_parent/2$ and $blue\_parent/2$ and additional definitions (omitted for
brevity) of the predicates $red\_child/2$, $blue\_child/2$, $parent/2$,
$child/2$. We give as metarules \emph{Identity}, \emph{Inverse}, \emph{Stack},
\emph{Queue} from Table \ref{tab:h22_metarules}, that match the clauses of the
target theory. The background knowledge and metarules suffice to reconstruct the
target theory, but mislabelled examples allow a correct hypothesis to be formed
only for the No Noise problem. 

\subsubsection{Coloured graph - results}
\label{subsubsec:Coloured graph - results}

Figures \ref{fig:redundant_bk} and \ref{fig:redundant_bk_time} plot the accuracy
and training time results of the Coloured graph experiment, respectively. In the
three ``noisy" datasets a correct hypothesis does not exist in $\mathcal{H}$ and
so Metagol's accuracy is that of the empty hypothesis (varying according to
mislabelled examples). Metagol tests a learned hypothesis against the negative
examples only once the hypothesis is completed, then backtracks to try a new
hypothesis if the test fails. This causes much backtracking in the False
Negatives dataset, so much so that Metagol exhausts the training time limit of
300 sec. for most of the experiment. Louise outperforms Metagol in all but the
No Noise dataset, although its performance fluctuates as the chance of
processing mislabelled examples increases with the size of the training
partition. In the No Noise dataset a short, correct hypothesis exists -the
target theory- and Metagol finds it earlier in the experiment than Louise. The
hypothesis space for this problem includes many over-general hypotheses formed
with predicates other than $ancestor/2$ which suffices to express the target
theory. Additional background predicates may be seen as, in a sense,
``redundant" and it is this redundancy that leads to Louise's reduced early
accuracy with No Noise.

We repeat the experiment with the redundant predicates removed, leaving
$ancestor/2$ as the only background predicate. Figures \ref{fig:no_redundant_bk}
and \ref{fig:no_redundant_bk_time} plot the accuracy and training time results,
respectively. The size of $\mathcal{H}$ is now reduced by several orders of
magnitude (see table \ref{tab:mil_problem_summary}). Metagol's predictive
accuracy remains unchanged but it can exhaustively search $\mathcal{H}$ and exit
with failure much more quickly in the ``noisy" datasets. Louise's accuracy
improves on the No Noise dataset but deteriorates in the False Negatives
dataset. Louise performs worse than the empty hypothesis in the Ambiguities
dataset, where the combination of mislabelled positive and negative examples
forces Algorithm \ref{alg:top_program} to form a Top program that entails  not
only few positive, but also many negative examples.

The results in this section support Experimental Hypothesis \ref{hyp:noise}.

\subsection{Experiment 3: M:tG Fragment} 
\label{subsec:Experiment 3: M:tG Fragment} 

\begin{table}[t]
	\centering
	\begin{tabular}{l}
		\textbf{M:tG Fragment} \\
		\toprule
		Positive examples \\
		\midrule
		$ability([destroy,target,artifact],[]).$ \\
		$ability([exile,all, \mlq Djinns \mrq],[]).$ \\
		$ability([exile,target, \mlq Hippogriff \mrq],[]).$ \\
		$ability([return,an,artifact,from,a,graveyard,to,its, \mlq owner\backslash's \mrq,hand],[]).$ \\
		$ability([return,target,planeswalker,to,its, \mlq owner\backslash's \mrq,hand],[]).$ \\
		$ability([return,all,creatures,from,your,graveyard,to,the,battlefield],[]).$ \\
		\midrule
		Background knowledge \\
		\midrule
		$ability \longrightarrow destroy\_verb, target\_permanent.$ \\
		$destroy\_verb \longrightarrow [destroy].$ \\
		$target\_permanent \longrightarrow target, permanent\_type.$ \\
		$target \longrightarrow [target].$ \\
		$permanent\_type \longrightarrow [artifact].$ \\
		$permanent\_type \longrightarrow [creature].$ \\
		$permanent\_type \longrightarrow [enchantment].$ \\
		$permanent\_type \longrightarrow [land].$ \\
		$permanent\_type \longrightarrow [planeswalker].$ \\
		\bottomrule
	\end{tabular}
\caption{M:tG fragment dataset: examples of positive example strings and
background knowledge comprised of grammar productions in Definite Clause
Grammars form. Tokens in square braces are terminals, other tokens are
nonterminals. ``$\longrightarrow$" can be read as ``expands to".}
\label{tab:mtg_fragment_dataset}
\end{table}

When each positive example in a MIL problem is entailed by exactly one clause in
$\Theta$, $\mathcal{H}$ ``collapses" to a single correct hypothesis.  This
permits us to test Experimental Hypothesis \ref{hyp:same_accurracy}.

\emph{Magic: the Gathering (M:tG)} is a Collectible Card Game played with cards
printed with instructions in a Controlled Natural Language (CNL) for which no
complete formal specification is published. We hand-craft a grammar in Definite
Clause Grammar form for a simple fragment of the M:tG CNL that includes only
expressions beginning with one of the three ``keyword actions" \emph{destroy},
\emph{exile} and \emph{return}. We manually extract the rules of the grammar
from two sources: a) examples of strings on cards and b) semi-formal
specifications of expressions provided in the game's rulebook \citep{WotC2018}.
Such specifications are provided for only a few expressions in the language,
most of which are pre-terminals denoting card types (e.g. $permanent\_type//0$
in table \ref{tab:mtg_fragment_dataset}). Each example string has a single parse
tree and so is entailed by exactly one rule in our grammar.

To set up a MIL problem for this dataset we generate all 1348 strings entailed
by our grammar to use as positive examples of the predicate $ability/2$ (the
start symbol of the grammar). We use the 60 nonterminals and pre-terminals in
our hand-crafted grammar as background knowledge and use \emph{Chain} as the
only metarule. The 36 productions of our grammar where the start symbol,
$ability/2$ is the nonterminal on the left-hand side are all instances of
\emph{Chain}, therefore \emph{Chain} is sufficient to construct a correct
representation of our grammar. Examples of the elements of the MIL problem for
this dataset are given in table \ref{tab:mtg_fragment_dataset}.

\subsubsection{M:tG Fragment - results}
\label{subsubsec:M:tG Fragment - results}

Figures \ref{fig:mtg_fragment_results} and \ref{fig:mtg_fragment_time} plot the
accuracy and training time results, respectively, of the M:tG Fragment
experiment. Louise and Metagol learn identical hypotheses (i.e. the Top program)
and their accuracy curves coincide. Louise is slightly faster for most of the
experiment but its training time ``spikes" towards the end of the experiment,
likely because of redundancy in the examples set that causes the same clauses to
be derived from different examples, multiple times\footnote{This inefficiency is
addressed in the current version of Louise by a variant of Algorithm
\ref{alg:top_program} that uses a coverset algorithm, discussion of which is
left for future work}. Metagol only learns a single clause from each example
thereby avoiding this duplication of effort. Even so Louise's training times
remain under 3.0 sec. for the entire experiment. The results of this experiment
confirm our Experimental Hypothesis \ref{hyp:same_accurracy}.

We note that the hypothesis learned by Metagol and Louise in this experiment is
exactly the target theory for the M:tG Fragment MIL problem and the size of this
target theory is 36 clauses, just over 7 times larger than any program learned
by Metagol previously reported in the MIL literature. This further supports
Experimental Hypothesis \ref{hyp:maximal_search}. When $\mathcal{H}$ is small,
even when $\Theta$ is large, Louise does not have a clear advantage over
Metagol.

\subsection{Discussion}
\label{subsec:Discussion}

The results in the previous sections show that Louise outperforms Metagol when
Metagol cannot find a correct hypothesis within the training time limit. This
is most evident in Experiment 1 (Figures \ref{fig:grid_world_results} and
\ref{fig:grid_world_time}) where both $\mathcal{H}$ and $\Theta$ are large and
Metagol's search is at its most expensive, and in the noisy datasets in
Experiment 2 (Figures \ref{fig:redundant_bk}, \ref{fig:no_redundant_bk},
\ref{fig:redundant_bk_time}, \ref{fig:no_redundant_bk_time}) where no correct
hypothesis exists in $\mathcal{H}$.

Metagol learns in two stages: first it finds a hypothesis, $H$, that is not
too-specific (i.e. $H \wedge B \models E^+$); then it tests $H$ against $E^-$.
If $H$ is over-general (i.e. if $H \wedge B \models e^- \in E^-$) Metagol
backtracks and searches for a new $H$. False positives in $E^+$ cause Metagol to
find over-general hypotheses that lead to much backtracking, increasing training
times early in the False Positives and Ambiguities experiments with large
$\mathcal{H}$ (Figure \ref{fig:redundant_bk_time}). Later in the same
experiments, the number of false positives sampled increases and the number of
not-too-specific hypotheses diminishes allowing Metagol to exit quickly with
failure. False negatives in $E^-$ cause many hypotheses to appear over-general
causing much backtracking in the False Negatives experiment with large
$\mathcal{H}$ (Figure \ref{fig:redundant_bk_time} False Negatives). In the
small-$\mathcal{H}$ experiments, $\mathcal{H}$ is small enough that Metagol's
search can exit quickly with failure (Figure \ref{fig:no_redundant_bk_time}).

Louise does not test hypotheses for generality and instead returns the
best-possible Top program without performing a search or backtracking so its
training times stay short with both large and small $\mathcal{H}$ (Figures
\ref{fig:redundant_bk_time}, \ref{fig:no_redundant_bk_time}) with small
fluctuations caused by redundancies in $E^+, E^-$. Louise's accuracy suffers
when $\mathcal{H}$ includes many over-general hypotheses because of irrelevant
background knowledge (Figure \ref{fig:redundant_bk} No Noise). However, Louise
can complete a learning attempt and return a result in situations where Metagol
continues to search for a very long time (Figures \ref{fig:grid_world_time},
\ref{fig:redundant_bk_time} False Negatives). These observations indicate that
Louise is better suited than Metagol to learning in large, complex problem
domains with classification noise.

%% file: 6_conclusions.tex
\section{Conclusions and future work}
\label{sec:conclusions}

\subsection{Conclusions} 
\label{subsec:Conclusions} 

We have shown that a costly search of the MIL hypothesis space, $\mathcal{H}$,
for a correct hypothesis can be replaced by the construction of a Top program,
$\top$, the set of clauses in all correct hypotheses, which is itself a correct
hypothesis that can be constructed without search, from a finite number of
examples and in polynomial time with Algorithm \ref{alg:top_program}. 

We have implemented Algorithm \ref{alg:top_program} in Prolog as the basis of a
new MIL system, called Louise, that learns by Top program construction and
reduction. We have compared Louise to the state-of-the-art search-based MIL
system, Metagol, and shown that Louise outperforms Metagol when the size of
$\mathcal{H}$ and the target theory, $\Theta$, are both large, because of
Metagol's exponential time complexity, or when the hypothesis space does not
include a correct hypothesis. The latter is the case e.g. when a MIL problem
includes classification noise and we have shown that Louise is more robust to
certain kinds of noise than Metagol. Louise does not have an advantage over
Metagol when $\mathcal{H}$ or $\Theta$ are small and we have found to our
surprise that Metagol can learn a hypothesis 7 times larger than any program
previously learned by Metagol, as reported in the MIL literature, when
$\mathcal{H}$ includes a single hypothesis which is, tautologically, $\top$.

\subsection{Future work}
\label{subsec:Future work}

\begin{table}[t]
	\centering
	\begin{tabular}{l}
		$\boldsymbol{a^nb^n}$ \textbf{grammar}\\
		\toprule
		$\mlq \$1 \mrq (x,y)\leftarrow \mlq S \mrq (x,z),\mlq B \mrq (z,y).$ \\
		$\mlq S \mrq (x,y)\leftarrow \mlq A \mrq (x,z),\mlq \$1 \mrq (z,y).$ \\
		$\mlq S \mrq (x,y)\leftarrow \mlq A \mrq (x,z),\mlq B \mrq (z,y).$ \\
		\bottomrule
	\end{tabular}
\caption{Definite Clause Grammar hypothesis for the $a^nb^n$ language learned
by Louise with Dynamic Learning. The definition of predicate $\mlq \$1 \mrq$ is
invented.}
\label{tab:anbn_predicate_invention}
\end{table}

An important limitation of our approach, demonstrated in section
\ref{subsubsec:Coloured graph - results}, is that Algorithm
\ref{alg:top_program} is forced to learn an over-general Top program when
$\mathcal{H}$ includes many over-general hypotheses and there are insufficient
negative examples to eliminate over-general clauses. In addition, Plotkin's
algorithm may not always remove clauses that are not logically redundant but
entail overlapping sets of examples. Louise implements two additional program
reduction procedures that address these limitations by selecting subsets of the
Top program that comprise correct hypotheses of minimal size (and with clauses
entailing non-overlapping sets of examples).

Louise is capable of predicate invention by recursive Top program construction
in an incremental learning setting named \emph{Dynamic Learning} (an example of
predicate invention in Louise's Dynamic Learning setting is listed in Table
\ref{tab:anbn_predicate_invention}). Finally, Louise implements a form of
\emph{examples invention} by semi-supervised learning similar to
\citep{Dumancic2019}. We have omitted discussion of these features for the sake
of brevity but plan to include them in upcoming work.

As a MIL system, Louise relies on the selection of relevant metarules, which is
currently left to user expertise. Selection of strong inductive biases by user
expertise (or intuition) is common in machine learning, e.g. in the selection
and careful fine-tuning of a neural network architecture, priors in Bayesian
learning, kernels in Support Vector Machines, etc. Previous work in the MIL
literature has addressed the issue of automatic selection of metarules, e.g.
\citep{Cropper2015} and \citep{Cropper2018}. Louise includes libraries for
metarule extraction from arbitrary Prolog programs (including background
knowledge definitions), as described in section \ref{subsec:A note on metarule
selection}; for metarule generation; and for metarule combination by unfolding.
Finally, predicate invention can effectively extend the set of metarules in a
MIL problem beyond those given initially by a user, as first noted in
\citep{Cropper2015} and investigated further in our upcoming work on the
Dynamic Learning setting. A more complete discussion of automatic selection of
metarules is left for future work.

The observation noted in section \ref{subsec:Experiment 3: M:tG Fragment} that
when each positive example is entailed by exactly one clause in the target
theory, the MIL hypothesis space includes a single program, merits further
theoretical and empirical investigation.

We have shown the existence of finite upper bounds on the numbers of examples
necessary for Top program construction with Algorithm \ref{alg:top_program},
but we have not derived sample complexity results. Previous work in the MIL
literature, e.g. \citep{Cropper2016}, has derived sample complexity results for
a search of $\mathcal{H}$ under PAC Learning assumptions \citep{Valiant1984}
and according to the Blumer Bound \citep{Blumer1987377}. Such results can also
be derived for Top program construction.

We have situated the Top program construction framework in the context of MIL
but a Top program should exist in any ILP setting. Such a more general
description of our framework remains to be done. Similarly, Top program
construction should be possible to implement in a different language, other than
Prolog, such as Answer Set Programming (ASP) etc. Indeed, MIL has also been
implemented in ASP, as \emph{hexmil} in \citep{Kaminski2018ExploitingAS} and
future work should compare our Prolog implementation of Louise against this MIL
implementation.

Finally, we are eager to test Louise's mettle on novel experimental
applications, particularly real-world applications in domains that have
traditionally proven hard for ILP because of the size of $\mathcal{H}$, as e.g.
in machine vision.

%% file: 7_acknowledgements.tex
\section{Acknowledgements}
\label{sec:Acknowledgements}

The first author acknowledges support from the UK's EPSRC for financial support
of her studentship. The second author acknowledges support from the UK's EPSRC
Human-Like Computing Network, for which he acts as director. We thank Lun Ai,
Wang-Zhou Dai and C\'{e}line Hocquette for reading and discussing early versions
of this paper and the anonymous reviewers for suggesting valuable improvements
to the paper.